\documentclass[a4paper,twoside]{article}

\usepackage{epsfig}
\usepackage{subcaption}
\usepackage{calc}
\usepackage{amssymb}
\usepackage{amstext}
\usepackage{amsmath}
\usepackage{amsthm}
\usepackage{multicol}
\usepackage{pslatex}
\usepackage{apalike}
\usepackage{algorithm2e}
\usepackage[bottom]{footmisc}
\usepackage{multirow}
\usepackage{comment}
\usepackage[table]{xcolor}
\usepackage{SCITEPRESS}     

\begin{document}

\title{A Comparative Study of 3D Person Detection: Sensor Modalities and Robustness in Diverse Indoor and Outdoor Environments}

\author{\authorname{Malaz Tamim*\orcidAuthor{0009-0002-9987-6645}, Andrea Matic-Flierl*\orcidAuthor{0000-0002-2179-0350} and Karsten Roscher\sup{}\orcidAuthor{0000-0002-9458-104X}}
\affiliation{\sup{}Fraunhofer Institute for Cognitive Systems IKS, Hansastr. 32, 80686 Munich, Germany}
\email{\{malaz.tamim, andrea.matic-flierl, karsten.roscher\}@iks.fraunhofer.de}
}


\abstract{Accurate 3D person detection is critical for safety in applications such as robotics, industrial monitoring, and surveillance. This work presents a systematic evaluation of 3D person detection using camera-only, LiDAR-only, and camera-LiDAR fusion. While most existing research focuses on autonomous driving, we explore detection performance and robustness in diverse indoor and outdoor scenes using the JRDB dataset. We compare three representative models — BEVDepth (camera), PointPillars (LiDAR), and DAL (camera–LiDAR fusion) — and analyze their behavior under varying occlusion and distance levels. Our results show that the fusion-based approach consistently outperforms single-modality models, particularly in challenging scenarios. We further investigate robustness against sensor corruptions and misalignments, revealing that while DAL offers improved resilience, it remains sensitive to sensor misalignment and certain LiDAR-based corruptions. In contrast, the camera-based BEVDepth model showed the lowest performance and was most affected by occlusion, distance, and noise. Our findings highlight the importance of utilizing sensor fusion for enhanced 3D person detection, while also underscoring the need for ongoing research to address the vulnerabilities inherent in these systems.}

\onecolumn \maketitle \normalsize \setcounter{footnote}{0} \vfill

\section{{\uppercase{introduction}}}

3D person detection is a key task in computer vision, attracting growing interest across various domains. While much of the existing research and model development has focused on automotive applications, there is clear potential for deploying these technologies in other areas, like robotics and industrial safety. Ensuring the safety of humans in these settings requires not only accurate detection but also robust performance under challenging conditions, like occlusion.

\renewcommand{\thefootnote}{}
\footnotetext{* Both authors contributed equally to this work.}
\renewcommand{\thefootnote}{\arabic{footnote}}

Modern 3D detection systems often use data from camera and LiDAR sensors. Cameras provide rich semantic information, enabling fine-grained classification and contextual understanding. LiDAR offers precise geometric data and benefits in terms of data privacy, as it does not capture identifiable visual features. Sensor fusion aims to combine the detection strengths of both modalities. 

However, adapting these systems to non-automotive domains presents a distinct set of challenges. A central issue is the domain gap: models trained on outdoor environments, such as roads and intersections, may not generalize well to datasets collected in different contexts. Indoor scenes or non-automotive outdoor environments often differ in terms of spatial layout, surface materials, lighting conditions, and the types of objects present. These variations can significantly influence the characteristics of both camera and LiDAR data, affecting how detection algorithms interpret and respond to the scene. As a result, performance may degrade when models are applied outside their original domain, highlighting the need for dedicated evaluation on more diverse datasets.

In safety-critical applications it is essential to evaluate not only the accuracy of pedestrian detection systems but also their reliability under diverse conditions. Real-world deployments often involve challenges such as sensor noise, calibration errors, and occlusion, all of which can significantly affect detection performance. Robustness to these factors is crucial to ensure consistent and trustworthy operation, especially when human safety is at risk. Several studies have investigated these aspects, particularly within the automotive domain.
This leaves open questions about how well current methods perform in other areas with different sensing conditions and scene characteristics.

To bridge this gap, we present a systematic evaluation of 3D person detection methods using camera, LiDAR, and their fusion, focusing on the JRDB dataset~\cite{martin-martin_jrdb_2023}, \cite{le_jrdb-panotrack_2024}, which offers a diverse set of indoor and outdoor scenes captured by a mobile robot on a university campus. We investigate how each modality behaves under varying conditions, such as occlusion, person distance, and sensor noise, thus providing new insights into the robustness and applicability of 3D detection systems beyond autonomous driving.

Our main contributions are the following:

\begin{itemize}
    \item We conduct a comparative analysis of 3D person detection using representative models for camera-only, LiDAR-only, and camera-LiDAR fusion approaches on diverse indoor and outdoor scenes, as contained in the JRDB dataset.
    \item We provide a comprehensive evaluation across modalities, examining 3D detection performance under varying conditions such as person distance and occlusion levels.
    \item We simulate various types of sensor noise and perform a systematic robustness analysis under these synthetic corruptions, assessing the impact on detection performance for each modality individually as well as the fused approach.
\end{itemize}

\section{{{\uppercase{Related work}}}}

\subsection{3D Object Detection}

Camera-based 3D detection methods are attractive due to their low cost and the rich semantic information in RGB images. However, estimating accurate depth from images alone remains challenging, especially for monocular approaches. 
In contrast, multi-view methods enhance spatial reasoning by utilizing multiple synchronized views. BEVDepth~\cite{bevdepth} advances this by integrating LiDAR-supervised depth estimation during training,  
achieving strong performance using only camera input at inference.

LiDAR-based methods provide accurate depth data, although their usage is often limited by the high cost of LiDAR sensors. These methods can be divided into four categories~\cite{song_robustness-aware_2024}: (1) View-based methods, (2) point-based methods, (3) Voxel-based methods, and (4) Point-voxel-based methods.
PointPillars~\cite{pointpillars} is a particularly efficient voxel-based approach: it organizes point clouds into vertical “pillars” and applies 2D convolutions to extract features, resulting in high speed and accuracy.

Sensor fusion methods combine the strengths of different modalities but increase system complexity, requiring precise calibration, synchronization, and more computational resources. A variety of approaches exist with differing fusion strategies. For example, 
BEVFusion~\cite{bevfusion} projects LiDAR and image features into a shared BEV representation. Building on this, DAL (Detecting As Labeling) decouples image features from the regression task, thereby reducing overfitting~\cite{dal}.

\subsection{3D person detection}

Several studies have systematically investigated the performance of different sensor modalities for 3D person detection, including effects from occlusion and varying distances. \cite{corral-soto_understanding_2020} analyze early fusion of RGB and LiDAR data, revealing modality-specific strengths depending on object distance and lighting. 
~\cite{semanticvoxels} introduce SemanticVoxels~\cite{semanticvoxels}, which fuses LiDAR data with semantic segmentation, and study its performance under varying occlusion levels.

These works are primarily situated in automotive contexts, but some recent efforts have begun to explore also other domains. \cite{blanch_lidar-assisted_2024} adapt automotive 3D detection models for video surveillance, but still rely on driving datasets. \cite{linder_cross-modal_2021} investigate human detection in industrial environments using a custom warehouse dataset. They propose a transfer learning strategy with RGB-D supervision to improve 3D LiDAR detection. 
Finally, \cite{jia_2d_2022} present a comprehensive comparison of state-of-the-art 2D and 3D LiDAR-based person detection methods on the JRDB dataset, highlighting trade-offs between accuracy, robustness, and efficiency within LiDAR modalities. Similar to their work, we also address person detection on the JRDB dataset, but focus on systematically comparing the performance and robustness between camera-only, LiDAR-only, and camera–LiDAR fusion approaches.

\subsection{Robustness of 3D Detection Models}

Robustness to sensor and environmental noise is essential for reliable 3D object detection in real-world applications. Recent studies have introduced automotive benchmarks such as KITTI-C, nuScenes-C, Waymo-C~\cite{dong_benchmarking_2023}, and nuScenes-R and Waymo-R~\cite{yu_benchmarking_2023}, simulating diverse corruption types to evaluate model stability. These works consistently show that multi-modal approaches outperform single-modality models on automotive datasets and that LiDAR-camera fusion offers the highest robustness, although it is still sensitive to LiDAR disruptions.
The review in~\cite{song_robustness-aware_2024} further emphasizes the importance of robustness alongside accuracy and latency. 

\vspace{0.3 cm}
\noindent Despite these contributions, systematic evaluations of sensor modality performance in indoor, non-automotive environments remain limited. In particular, there is a lack of studies that rigorously assess the reliability and robustness of camera-only, LiDAR-only, and camera-LiDAR fusion approaches for such domains, especially under challenging conditions such as occlusion and sensor noise. 
\begin{figure*}[!ht]
  \centering
   \includegraphics[width=\textwidth]{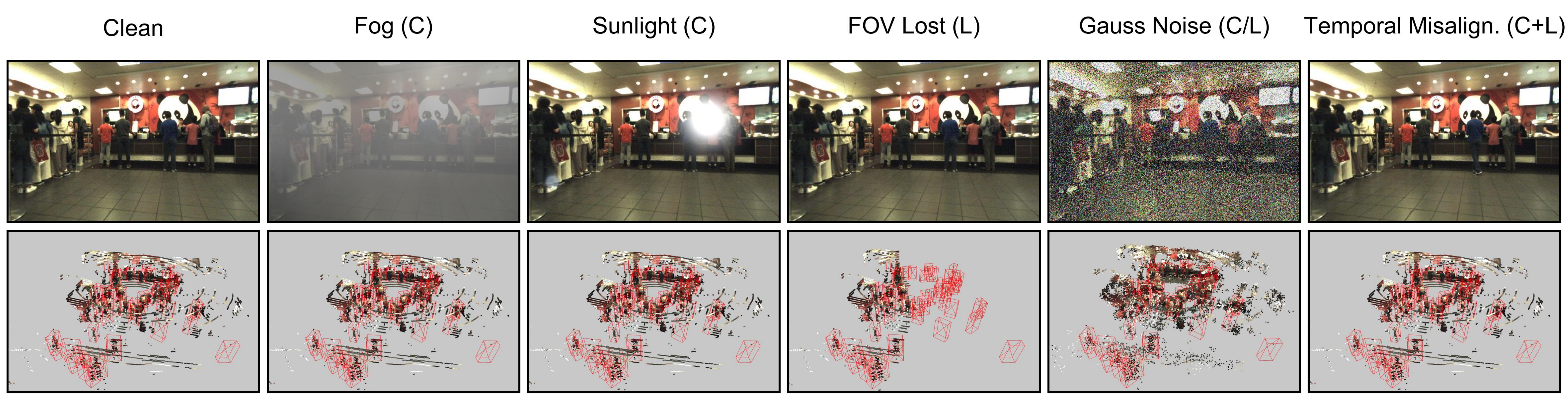}
    \caption{\textbf{Visualization of typical camera and LiDAR corruptions.} The top row shows one of the five camera views of the multi-view setup in JRDB; the bottom row shows the corresponding 360° LiDAR point cloud with ground-truth boxes in red. Modalities are annotated as C (camera) and L (LiDAR).}

  \label{fig:corruptions}
\end{figure*}

\vspace{0.1cm}

\section{\uppercase{corruptions}}

\label{sec:corruptions}

To assess the effects of various noise sources on 3D person detection performance, we apply synthetically generated corruptions as designed by~\cite{dong_benchmarking_2023}. These corruptions are artificially added to the dataset to emulate realistic sensor imperfections and environmental challenges commonly observed in autonomous driving. The corruptions are derived from empirical observations of real-world failure modes in camera and LiDAR systems, making them representative of practical robustness issues encountered in deployment. 
Among these, we examine selected corruptions, focusing on sensor noise and calibration errors to study the impact of each modality on detection results. Each corruption type is simulated with three severity levels, corresponding to the 1st, 3rd, and 5th levels defined in~\cite{dong_benchmarking_2023}. Representative examples on camera and LiDAR images are illustrated in Figure~\ref{fig:corruptions}. The corruptions, along with their technical implementation, will be explained in the following.

\subsection{Sensor-Level Corruptions}

Sensor noise is caused by internal and external factors, like sensor vibrations, reflective materials and lighting conditions. 
Gaussian noise is characterized by random fluctuations in the data. For LiDAR, this represents ranging inaccuracies. In our studies, it is added to all points with severities of {0.02m, 0.06m, 0.10m}. 
In camera images, Gaussian noise originates from camera defects or low lighting conditions. It is modeled through image augmentation.

The other sensor corruptions are applied only to LiDAR. Cutout simulates scenarios where portions of the LiDAR data are missing, e.g. due to surface reflectivity issues. This corruption is implemented by dividing the point-cloud data in a frame into 50 groups of points and then randomly dropping 2, 5 or 10 of these groups. 
Crosstalk noise occurs when multiple LiDAR sensors operate in close proximity.
It is modeled by applying Gaussian noise to a subset of points. The ratio of selected points for crosstalk is set to {0.4\%, 1.2\%, 2\%}. 
Density decrease describes the scenario where the density of LiDAR points is reduced. This is implemented by randomly deleting a specified percentage of points from a LiDAR frame, set to {6\%, 18\%, 30\%}.
Field-of-view (FOV) loss represents situations where the field of view of the sensor is restricted, e.g. due to occlusions. This is simulated by removing groups of points in specific angular ranges. Following angular ranges are considered for the three severity levels: {(-75°, 75°), (-45°, 45°), (-105°, 105°)}.

\subsection{Sensor Misalignment}

We evaluate two types of sensor misalignment. Spatial misalignment occurs when the relative positions of LiDAR and camera sensors change over time, e.g. due to sensor vibrations. This is modeled by adding Gaussian noise to the calibration matrices, with noise levels set to {0.02, 0.06, 0.10} for the rotation matrix and {0.002, 0.006, 0.010} for the translation matrix.

Temporal misalignment arises when there is a delay in the data acquisition from different sensors, leading to inconsistencies in the camera and LiDAR frames used for detection. We simulate temporal misalignment by using outdated data from one sensor while keeping the other sensor's data current. These delays are modeled using fixed offsets of 2, 6, and 10 frames. We consider two implementations for this corruption: In one case we delay the camera images and in the other one the LiDAR frames.

\subsection{Weather}

While our primary focus is on sensor noise and misalignment, we also include two weather corruptions: fog and sunlight. Since the sensor corruptions contain only a single camera-level corruption, namely Gaussian noise, the weather corruptions serve as a valuable approach to study more noise effects on camera level. In contrast to~\cite{dong_benchmarking_2023}, we model weather effects only for camera and not for LiDAR. 

Fog, though less relevant in indoor environments, serves as an analog for effects of smoke or other particulate matter that can impair visibility. It is simulated by applying a gray mask layer to the images, with opacity levels of {10\%, 30\%, 50\%}. Sunlight is modeled by adjusting brightness and contrast in the images. 
\vspace{0.1cm}

\section{{\uppercase{Experimental setup}}}

This section describes the JRDB dataset, along with the training configuration and evaluation metrics used in our studies.

\subsection{JRDB Dataset}

The JackRabbot Dataset (JRDB)~\cite{martin-martin_jrdb_2023},~\cite{le_jrdb-panotrack_2024} is a multimodal dataset collected on a university campus using a social mobile manipulator. It features various indoor and outdoor scenes and provides 360° coverage from RGB and LiDAR sensors. The dataset consists of 54 sequences with a total length of 64 minutes. Each sequence is annotated with over 2.4 million 2D bounding boxes and associated 3D cuboids. The recorded scenes capture a variety of human postures across different crowd densities. Additionally, the annotations reflect several levels of person occlusion, categorized as fully visible, mostly visible, severely occluded, and fully occluded. In all experiments, we use the official JRDB 2022 split with 21,704 samples for training, 6,189 for validation, and 27,661 for testing, respectively.

\subsection{Training Setup}

In this work, we use BEVDepth, PointPillars and DAL as our baseline models for the three modalities (C, L and C+L, respectively). We build on the official implementations of all three models, introducing minimal modifications to image preprocessing and hyperparameter tuning to adapt them to JRDB and single-class person detection. All experiments are conducted 
with a global batch size of 16 on two NVIDIA L40 GPUs. The following subsections detail the model-specific configurations and settings used in our setup.

\subsubsection{BEVDepth}
We adopt the BEVDepth~\cite{bevdepth} framework with a Lift–Splat–Shoot view transformer that discretizes depth from $0$ to $50\,\mathrm{m}$ in $0.5\,\mathrm{m}$ steps and projects multi-view image features into a BEV representation. All camera images are undistorted to remove lens distortion effects, and we use the full $480\times 752$ resolution from five synchronized views, which together provide $360^\circ$ coverage. The BEV grid spans $[-40,40]\,\mathrm{m}$ in $x$ and $y$ with a $0.3125\,\mathrm{m}$ cell size, resulting in a $256\times 256$ grid. Detection is configured for a single class (person). Training runs for $20$ epochs with AdamW (learning rate $5\times10^{-5}$, weight decay $0.01$) under a step schedule with linear warm-up.

\subsubsection{PointPillars}

For LiDAR-only detection we employ the PointPillars~\cite{pointpillars} architecture. Raw point clouds are voxelized into vertical pillars with a size of $0.16 \times 0.16 \times 4.0\,\mathrm{m}$ within the range $[-39.68, -39.68, -2]\,\mathrm{m}$ to $[39.68, 39.68, 2]\,\mathrm{m}$. 
Training is carried out for 20 epochs using the AdamW optimizer (learning rate $1\times10^{-3}$, weight decay $0.01$) with a cyclic learning rate and momentum schedule.

\subsubsection{DAL}

The network architecture is identical to the original DAL design~\cite{dal}.  
The main differences lie in data preprocessing and training hyperparameters, which we adapt to the JRDB dataset and our hardware setup.

Similar to BEVDepth, all camera images are undistorted, and we keep the full sensor resolution as network input in order to preserve fine-grained details. Five synchronized camera views are fused with LiDAR point clouds covering the 3D range $[-40,\,-40,\,-2]\,\mathrm{m}$ to $[40,\,40,\,2]\,\mathrm{m}$.
Depth is discretized from $0\,\mathrm{m}$ to $50\,\mathrm{m}$ in $0.5\,\mathrm{m}$ steps, and the BEV grid spans $256 \times 256$ cells, corresponding to the $[-40,40]\,\mathrm{m}$ x--y range with $0.3125\,\mathrm{m}$ cell size. The vertical direction $[-2,2]\,\mathrm{m}$ is divided into four bins. LiDAR points are voxelized at voxel size of $0.05 \times 0.05 \times 0.2\,\mathrm{m}$.
For training, we run 15 epochs with a cyclic learning rate schedule and the AdamW optimizer (learning rate $5\times10^{-5}$, weight decay $0.01$).

\subsection{Evaluation Metrics}

We evaluate the 3D detection performance according to the JRDB benchmark using average precision (AP)~\cite{martin-martin_jrdb_2023}. Two thresholds for the 3D bounding boxes are considered: an Intersection over Union (IoU) threshold of 0.3 and a stricter threshold of 0.5. 
Following the official JRDB evaluation protocol, only those 3D bounding boxes are included which encompass more than 10 points and are located within 25 meters to the robot. 
\section{\uppercase{Results}}

In this section, we present results for the camera-only BEVDepth (C), LiDAR-only PointPillars (L), and DAL, which fuses camera and LiDAR data (C+L). First, we compare the performance of the different modalities, then analyze performance across distance and occlusion levels, followed by a robustness study.


\subsection{3D Detection Performance}
\label{sec:nominal_performance}

Table~\ref{tab:jrdb_val_results} and Table~\ref{tab:jrdb_test_results} show the AP values obtained with BEVDepth, PointPillars and DAL for the JRDB validation and test set, respectively. The fusion-based DAL consistently outperforms the single-modality models across both IoU thresholds, with an AP$_{0.3}$ of 73.18\% and AP$_{0.5}$ of 24.73\% on the JRDB test set. BEVDepth yields the lowest performance, which is also expected for the 3D detection performance of a camera-based approach. Across all models, we observe a notable drop in performance from AP${0.3}$ to AP${0.5}$, reflecting the difficulty of achieving high spatial alignment between predicted and ground truth 3D bounding boxes under stricter IoU constraints.

\begin{table}[ht]
\centering
\caption{3D detection performance of BEVDepth, PointPillars, and DAL on the JRDB 2022 validation set.}

\label{tab:jrdb_val_results}
\begin{tabular}{lccc}
\hline
\textbf{Method} & \textbf{Modality} & \textbf{AP$_{0.3}$} & \textbf{AP$_{0.5}$} \\
\hline
BEVDepth         & C   & 30.06 & 3.21 \\
PointPillars & L   & 51.47 & 17.97 \\
DAL              & C+L & 64.56 & 21.60 \\
\hline
\end{tabular}
\end{table}

\begin{table}[ht]
\centering
\caption{3D detection performance of BEVDepth, PointPillars, and DAL on the JRDB 2022 test set.}

\label{tab:jrdb_test_results}
\begin{tabular}{lccc}
\hline
\textbf{Methods} & \textbf{Modality} & \textbf{AP$_{0.3}$} & \textbf{AP$_{0.5}$} \\
\hline
BEVDepth         & C   & 24.84  & 2.70 \\
PointPillars     & L   & 58.08 & 17.56 \\
DAL              & C+L & 73.18 & 24.73 \\
\hline
\end{tabular}
\end{table}

While our implementations do not aim to achieve state-of-the-art performance, they provide a stable and representative baseline for the studies in the following sections. For context, the recently released DCCLA model~\cite{dccla} achieves an AP$_{0.3}$ of 76.28\% on the JRDB 2022 test set, representing the current upper bound on this dataset. However, BEVDepth, PointPillars and DAL were deliberately chosen as a compromise between established performance and reproducibility, with only minimal hyperparameter tuning.

\subsection{Varying Distance and Occlusion}

To assess the robustness of person detection models under real-world conditions, we conducted a comprehensive evaluation on the JRDB validation set, focusing on varying levels of distance, occlusion, and their combined effects. 

\subsubsection{Distance-Based Evaluation}

We analyzed detection performance across three distance categories: near (0–3 meters), mid (3–7 meters), and far (7–25 meters). These were selected such that each category contains approximately one third of the ground truth samples in the JRDB validation set.

\begin{figure}[!h]
  \centering
  \includegraphics[width=0.38\textwidth]{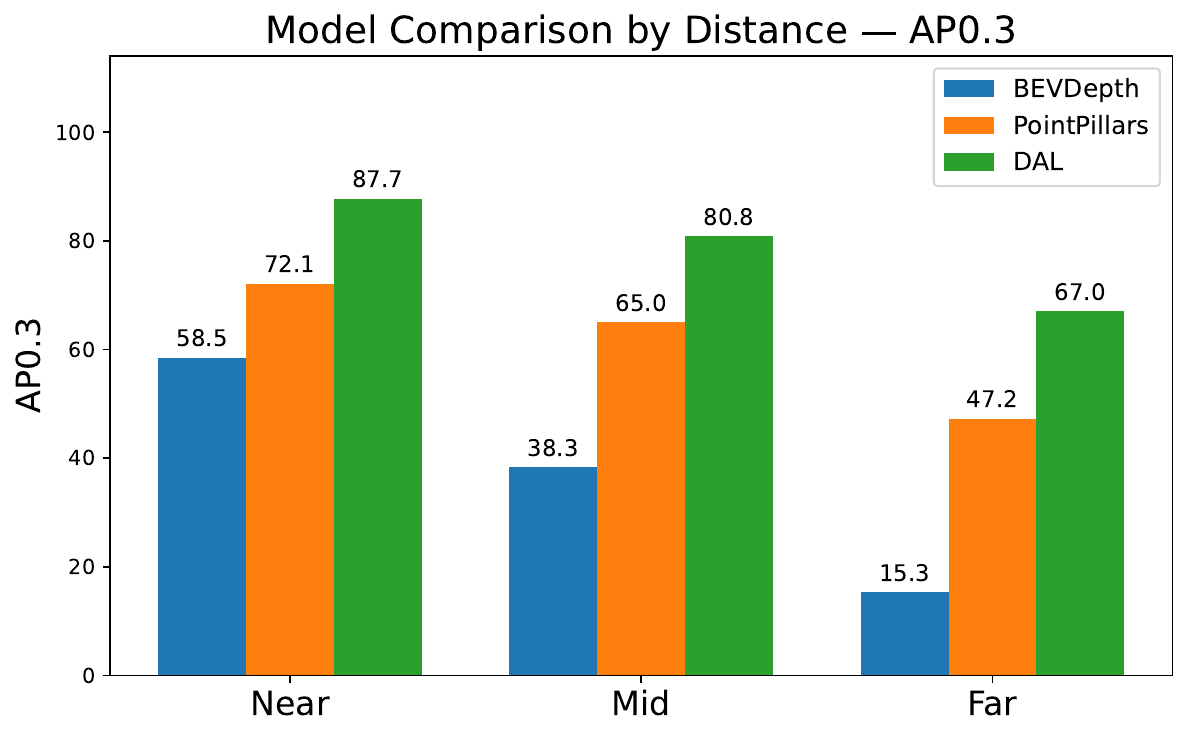}
  \includegraphics[width=0.38\textwidth]{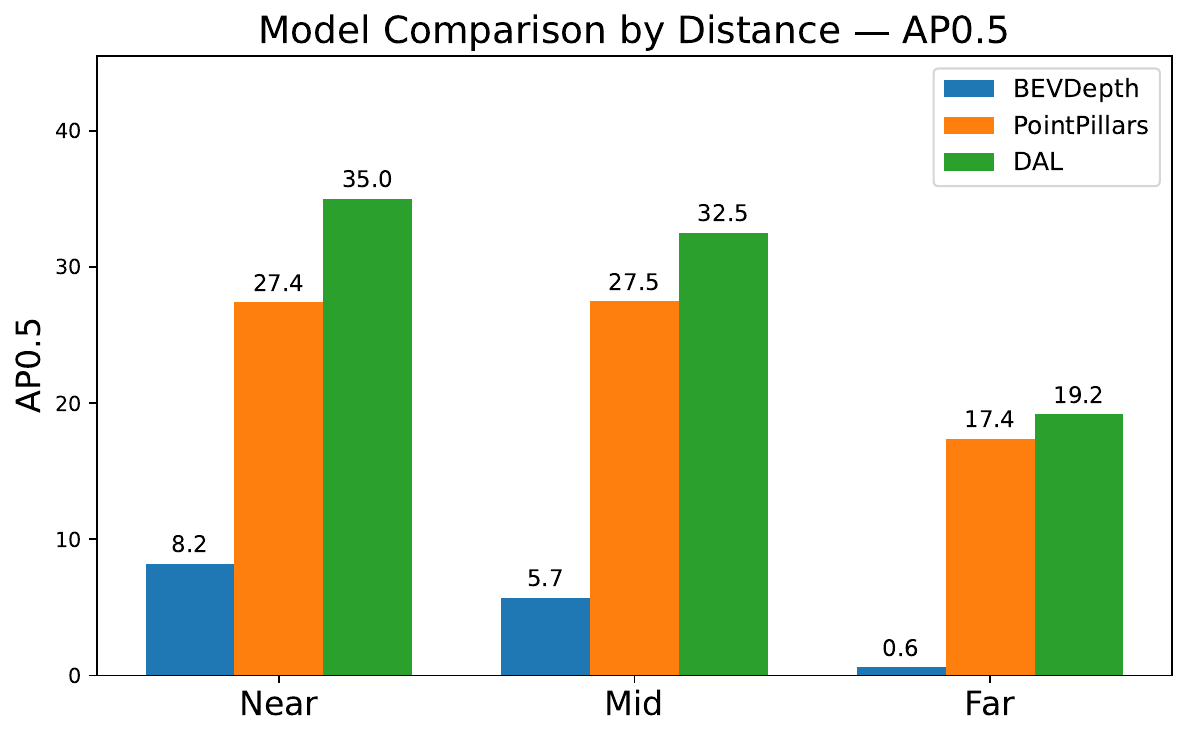}   
  \caption{Comparison of AP$_{0.3}$ (top) and AP$_{0.5}$ (bottom) by distance categories: near, mid, and far for BEVDepth, PointPillars, and DAL. }
  \label{fig:distance}
\end{figure}

Figure~\ref{fig:distance} shows the evaluation results using AP$_{0.3}$ and AP$_{0.5}$ across the distance categories. 
As expected, detection accuracy declined with increasing distance. BEVDepth exhibited the steepest drop, with a relative decrease of about 74\% in AP$_{0.3}$ from near to far (from 58.50\% to 15.35\%), while PointPillars and DAL showed moderate relative declines of about 24\% and 34\%, respectively. This underscores the relative resilience of fusion-based, and especially LiDAR approaches for long-range detection.
For AP$_{0.5}$, the performance gap between BEVDepth and the other models widened significantly compared to AP$_{0.3}$. In the far range, BEVDepth dropped to an AP$_{0.5}$ of just 0.64\%. PointPillars and DAL maintained comparable performance with 17.43\% and 19.19\%, respectively.

\subsubsection{Occlusion-Based Evaluation} 

\begin{figure}[!h]
  \centering
  \includegraphics[width=0.38\textwidth]{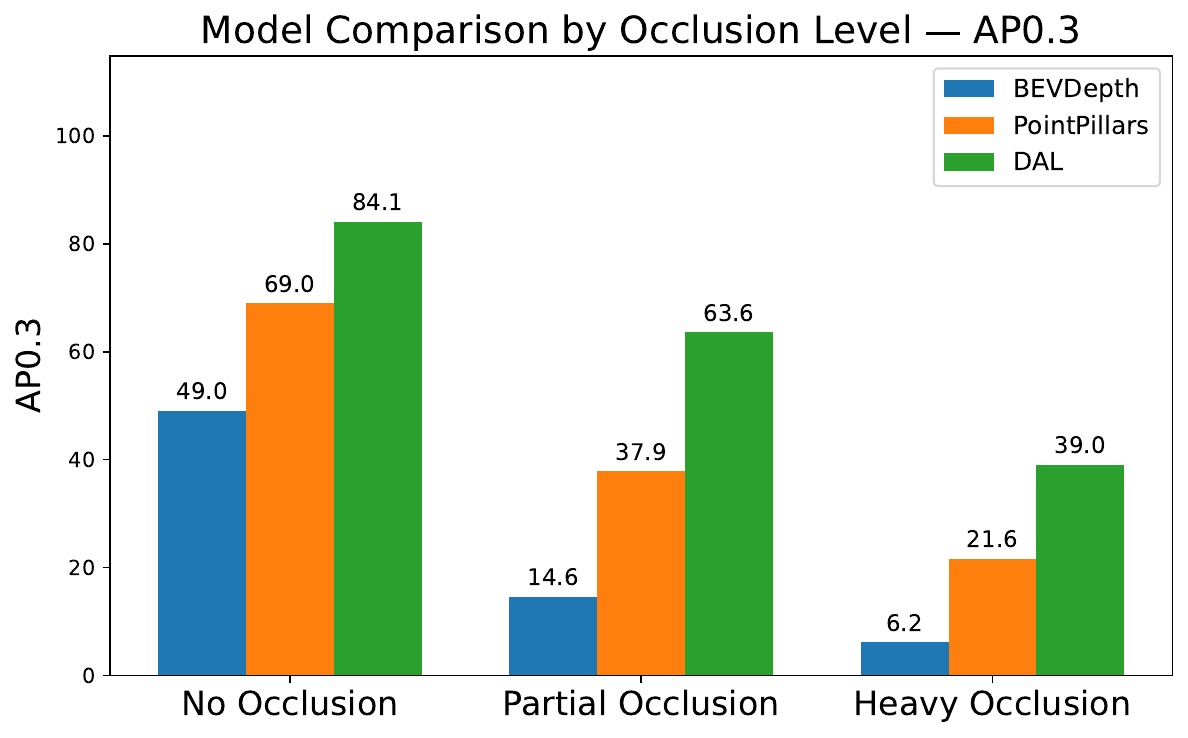}
  \includegraphics[width=0.38\textwidth]{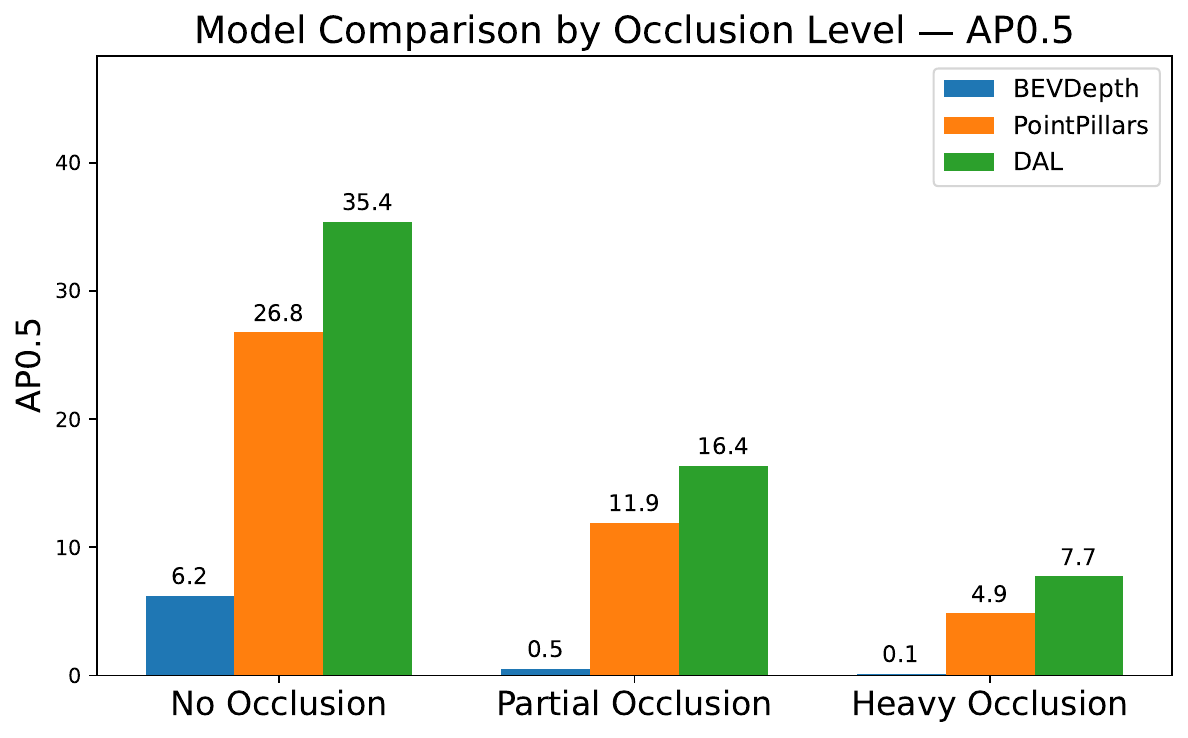}  
  \caption{Comparison of AP$_{0.3}$ (top) and AP$_{0.5}$  (bottom) across unoccluded, partially occluded, and heavily occluded categories for BEVDepth, PointPillars, and DAL. }
  \label{fig:occlusion}
\end{figure}

We next evaluated detection performance under varying occlusion levels, using the JRDB annotations to separate the ground-truth samples into no occlusion, partial occlusion, and heavy occlusion categories. 
Figure~\ref{fig:occlusion} shows the evaluation results across different occlusion levels. 
Similar to distance, performance decreased with increasing occlusion. For BEVDepth, performance dropped substantially, with a  relative reduction of 87\% with respect to the no-occlusion scenario. PointPillars and DAL were more robust, with relative drops of 69\% and 54\%, respectively, highlighting particularly the stability of the fusion-based DAL model.

For AP$_{0.5}$, the performance difference between BEVDepth and the other two models is much more pronounced than for AP$_{0.3}$. Overall, BEVDepth showed significant limitations, with AP$_{0.5}$ falling below 1\% in both partial and heavy occlusion scenarios. DAL gave the highest performance, achieving an AP$_{0.5}$ of 7.75\% for heavy occlusion. 

\subsubsection{Combined Evaluation} 

\begin{figure*}[!h]
  \centering
  \includegraphics[width=\textwidth]{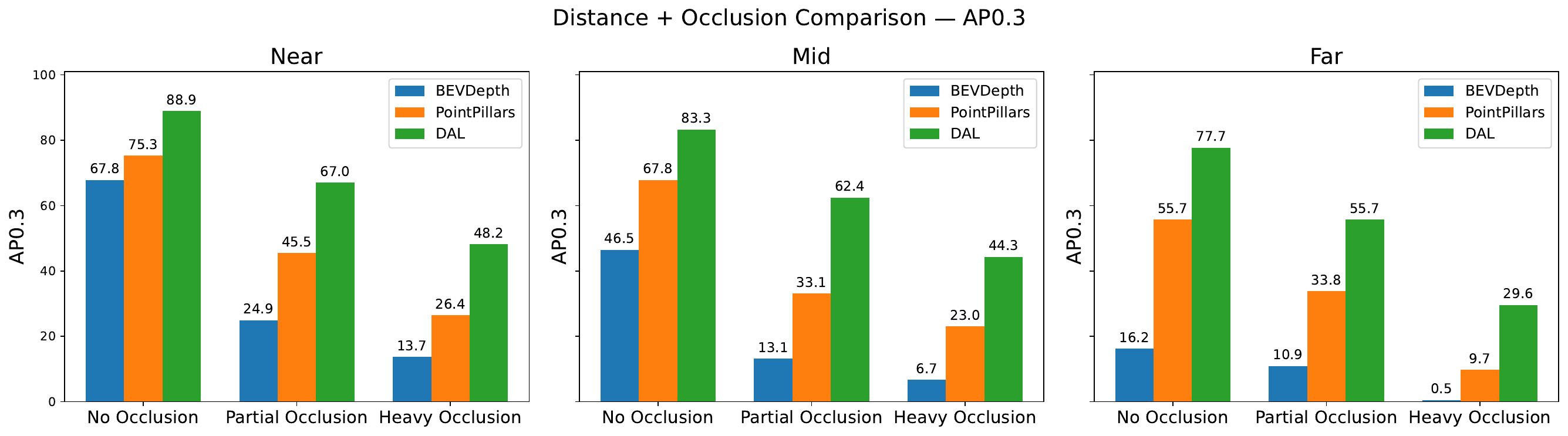}
  \includegraphics[width=\textwidth]{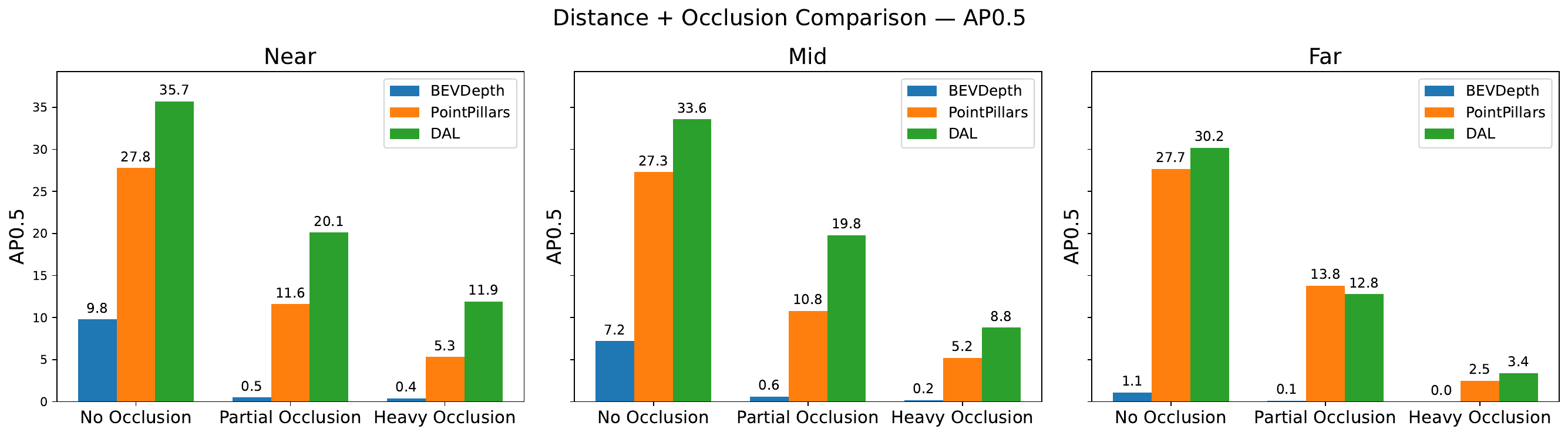}   
  \caption{Comparison of AP$_{0.3}$ (top) and AP$_{0.5}$ (bottom) across combined distance and occlusion categories for BEVDepth, PointPillars, and DAL. 
  }

  \label{fig:dist-occ}
\end{figure*}

We also examined the combined effects of distance and occlusion, which is illustrated in Figure~\ref{fig:dist-occ}. In the most favorable condition (no occlusion and near distance) all models achieved their highest AP$_{0.3}$: 67.76\% for BEVDepth, 75.28\% for PointPillars, and 88.86\% for DAL. However, performance deteriorated sharply under more challenging conditions. In the far and heavily occluded category, BEVDepth dropped to just 0.55\%, while DAL still maintained a relatively strong performance of 29.61\%.

At AP$_{0.5}$, BEVDepth consistently failed across all distance ranges under partial and heavy occlusion, with AP$_{0.5}$ values below 1\%. DAL outperformed PointPillars in the near and mid range, while in the far range their performance was quite similar. Notably, under partial occlusion at far distances, PointPillars slightly surpassed DAL. In the most difficult scenario (far distance with heavy occlusion) DAL and PointPillars achieved an AP$_{0.5}$ of 3.38\% and 2.50\%, respectively, while BEVDepth remained ineffective.

\subsection{Robustness Studies}

To assess the 3D detection robustness, we apply the corruptions of Section~\ref{sec:corruptions} to the JRDB validation set and evaluate the model performance. The evaluation was repeated for the three severity levels using the same models trained on the undistorted JRDB training set. Table~\ref{tab:corruptions} summarizes the results for AP$_{0.3}$.

\begin{table}[h]
    \centering
    \caption{Robustness study with AP$_{0.3}$ on the JRDB validation set across LiDAR (L), camera (C), and cross-modal (C+L) corruptions at severities 1–3. PointPillars (L), BEVDepth (C), and DAL (C+L) are reported; “None” denotes the uncorrupted baseline, while dashes denote corruptions that are not applicable to a given model.}
    \resizebox{\linewidth}{!}{
    \begin{tabular}{|l|c|c|c|c|}
        \hline
         \multirow{2}{*}{\textbf{Corruption}} & \multirow{2}{*}{\textbf{Level}} & \textbf{PointPillars} & \textbf{BEVDepth} & \textbf{DAL } \\
         & & \textbf{(L)} & \textbf{(C)} & \textbf{(C+L)} \\        
        \hline
       \rowcolor{gray!20} None & - & 51.47 & 30.06 & 64.27 \\
        \hline 
        \multirow{3}{*}{Cutout (L)} & 1 & 42.15 & - & 63.30 \\
        & 2 & 39.68 & - & 61.55 \\
        & 3 & 36.71 & - & 55.78 \\
        \hline
        \multirow{3}{*}{FOV Loss (L)} & 1 & 27.11 & - & 41.41 \\
        & 2 & 17.85 & - & 29.27 \\
        & 3 & 9.96 & - & 16.67 \\
        \hline
       {Density} & 1 & 42.77 & - & 64.27 \\
        Decrease & 2 & 42.12 & - & 64.17 \\
        (L) & 3 & 41.18 & - & 63.93 \\
        \hline
        Gaussian  & 1 & 43.33 & - & 62.98 \\
        Noise & 2 & 40.16 & - & 27.93 \\
        (L) & 3 & 29.39 & - & 12.80 \\
        \hline
        LiDAR & 1 & 27.44 & - & 63.65 \\
        Crosstalk & 2 & 13.28 & - & 62.37 \\
        (L) & 3 & 8.09 & - & 60.94 \\
        \hline
        \hline
        Gaussian & 1 & - & 9.09 & 64.27 \\
        Noise & 2 & - & 4.55 & 64.23 \\
        (C) & 3 & - & 1.82 & 64.21 \\
        \hline
        \multirow{3}{*}{Sunlight (C)} & 1 & - & 27.49 & 64.28 \\
        & 2 & - & 27.03 & 64.28 \\
        & 3 & - & 26.03 & 64.26 \\
        \hline
        \multirow{3}{*}{Fog (C)} & 1 & - & 19.18 & 64.12 \\
        & 2 & - & 12.55 & 64.10 \\
        & 3 & - & 9.09 & 64.01 \\
        \hline
        \hline
        Temporal & 1 & - & - & 64.25 \\
        Misalignment  & 2 & - & - & 64.21 \\
        Camera (C+L) & 3 & - & - & 64.18 \\
        \hline
        Temporal & 1 & - & - & 54.23 \\
        Misalignment & 2 & - & - & 31.93 \\
        LiDAR (C+L) & 3 & - & - & 20.78 \\
        \hline
        Spatial & 1 & - & - & 64.40 \\
        Misalignment & 2 & - & - & 64.24 \\
        (C+L) & 3 & - & - & 64.01 \\
        \hline
    \end{tabular}
    }
    \label{tab:corruptions}
\end{table}

\subsubsection{LiDAR-Based Corruptions} 

The LiDAR-based corruptions only influence PointPillars and DAL. Cutout led to a moderate performance decrease across the severity levels,
while a more pronounced degradation was observed under FOV loss, where the AP$_{0.3}$ score of PointPillars and DAL fell to 9.96\% and 16.67\%, respectively, for severity 3. This suggests that both models rely heavily on spatial coverage of the LiDAR sensor. In contrast, density reduction had minimal impact on DAL, with AP$_{0.3}$ consistently above 63\%. Surprisingly, PointPillars yielded an AP$_{0.3}$ of about 41-42\% across all severity levels, while the performance on the uncorrupted validation set is considerably higher (51.47\%). This is different to the observations in \cite{dong_benchmarking_2023}, where the LiDAR-based models suffered only minimal performance loss under density decrease. Two important differences between our and their setup are the different dataset domain and that we consider only one-class (person) detection. However, more studies would be necessary to understand if this causes the described behavior.

Under Gaussian noise, PointPillars showed a gradual decline from 43.33\% to 29.39\%. DAL, however, exhibited a sharp drop from 62.98\% to just 12.80\% at severity level 3. Interestingly, DAL initially outperformed PointPillars at low noise levels, but its performance deteriorated more rapidly. 
LiDAR crosstalk also reduced PointPillars’ performance significantly, with an AP$_{0.3}$ of 8.09\% at severity level 3, while DAL remained largely unaffected. 

\subsubsection{Camera-Based Corruptions} 

For BEVDepth, the camera-based corruptions had a severe impact. Gaussian noise led to a drastic performance drop, with AP$_{0.3}$ falling to just 1.82\% at severity level 3. Fog also degraded performance substantially. 
In contrast, sunlight had a relatively minor effect. BEVDepth’s performance decreased from 30.06\% (undistorted) to 26.03\% at severity level 3. The small variation across the three severity levels suggests that the corruption’s impact depends more on whether the sun is present in the image than on its intensity. This could be due to local brightness effects and glare affecting specific regions of the image.

DAL’s performance remained nearly constant across all camera-based corruptions, with AP$_{0.3}$ consistently around 64\%. This can be attributed to DAL’s architecture, where image features are excluded from the 3D bounding box regression task. As a result, localization is primarily driven by LiDAR input, making DAL largely resistant to camera distortions. Future work should study models like BEVFusion on JRDB, as it uses image features in the regression task.

\subsubsection{Sensor Misalignment} 

Sensor misalignment affects only fusion models due to their reliance on accurate cross-modal calibration. Temporal misalignment of the camera (i.e., delayed image frames) and spatial misalignment had minimal impact, with AP$_{0.3}$ remaining above 64\% across all severity levels. This again reflects the limited role of image features in DAL’s localization process.
However, temporal misalignment of LiDAR frames caused a significant performance drop down to an AP$_{0.3}$ of 20.78\%. This highlights the importance of temporal consistency in LiDAR data for accurate 3D detection.

\subsubsection{AP$_{0.5}$ Observations} 

While AP$_{0.3}$ revealed clear trends in robustness across corruption types and severity levels, AP$_{0.5}$ results were more erratic. No consistent patterns emerged, and performance fluctuated significantly. This suggests that the stricter IoU threshold may be too sensitive to small localization errors, making it less reliable for robustness evaluation in 3D detection tasks.

\section{\uppercase{Conclusion}}

We conducted a systematic evaluation of 3D person detection on the JRDB dataset using three representative models: BEVDepth (camera-only), PointPillars (LiDAR-only), and DAL (camera-LiDAR fusion). 
DAL outperforms single-modality approaches, especially under challenging conditions like occlusion, long-range detection, and sensor noise. While it offers superior performance and robustness, it remains sensitive to sensor misalignment and certain LiDAR noise. In contrast, the camera-based BEVDepth showed the lowest performance and was most affected by these challenges.
These findings provide a foundation for understanding 3D person detection in domains beyond autonomous driving, such as indoor environments, and represent a promising step towards developing reliable 3D person detection systems for applications like industrial safety monitoring. In future work, it will be important to study additional models, such as BEVFusion, and to evaluate performance on a wider range of datasets. This will help to further generalize the findings and identify model strengths and limitations across diverse real-world scenarios.

\section*{\uppercase{Acknowledgements}}
We acknowledge the use of Undermind for literature review support, as well as the Fraunhofer-internal FHGenie and Microsoft 365 Copilot for editing and writing assistance, such as phrasing improvements.

\bibliographystyle{apalike}
{\small
\bibliography{3D_robustness_paper}}

\begin{thebibliography}{}

\bibitem[Blanch et~al., 2024]{blanch_lidar-assisted_2024}
Blanch, M.~R., Li, Z., Escalera, S., and Nasrollahi, K. (2024).
\newblock {LiDAR}-{Assisted} {3D} {Human} {Detection} for {Video}
  {Surveillance}.
\newblock In {\em 2024 {IEEE}/{CVF} {Winter} {Conference} on {Applications} of
  {Computer} {Vision} {Workshops} ({WACVW})}, pages 123--131.

\bibitem[Corral-Soto and Bingbing, 2020]{corral-soto_understanding_2020}
Corral-Soto, E.~R. and Bingbing, L. (2020).
\newblock Understanding {Strengths} and {Weaknesses} of {Complementary}
  {Sensor} {Modalities} in {Early} {Fusion} for {Object} {Detection}.
\newblock In {\em 2020 {IEEE} {Intelligent} {Vehicles} {Symposium} ({IV})},
  pages 1785--1792.

\bibitem[Dong et~al., 2023]{dong_benchmarking_2023}
Dong, Y., Kang, C., Zhang, J., Zhu, Z., Wang, Y., Yang, X., Su, H., Wei, X.,
  and Zhu, J. (2023).
\newblock Benchmarking {Robustness} of {3D} {Object} {Detection} to {Common}
  {Corruptions} in {Autonomous} {Driving}.
\newblock In {\em 2023 {IEEE}/{CVF} {Conference} on {Computer} {Vision} and
  {Pattern} {Recognition} ({CVPR})}, pages 1022--1032.

\bibitem[Fei et~al., 2020]{semanticvoxels}
Fei, J., Chen, W., Heidenreich, P., Wirges, S., and Stiller, C. (2020).
\newblock {SemanticVoxels}: {Sequential} {Fusion} for {3D} {Pedestrian}
  {Detection} using {LiDAR} {Point} {Cloud} and {Semantic} {Segmentation}.
\newblock In {\em 2020 {IEEE} {International} {Conference} on {Multisensor}
  {Fusion} and {Integration} for {Intelligent} {Systems} ({MFI})}, pages
  185--190.

\bibitem[Guang et~al., 2025]{dccla}
Guang, J., Wu, S., Hu, Z., Zhang, Q., Wu, P., and Liu, J. (2025).
\newblock Dccla: Dense cross connections with linear attention for lidar-based
  3d pedestrian detection.
\newblock {\em IEEE Transactions on Circuits and Systems for Video Technology},
  35(5):4535--4548.

\bibitem[Huang et~al., 2025]{dal}
Huang, J., Ye, Y., Liang, Z., Shan, Y., and Du, D. (2025).
\newblock Detecting as {Labeling}: {Rethinking} {LiDAR}-{Camera} {Fusion} in
  {3D} {Object} {Detection}.
\newblock In {\em Computer {Vision} – {ECCV} 2024}, volume 15080, pages
  439--455. Springer Nature Switzerland, Cham.

\bibitem[Jia et~al., 2022]{jia_2d_2022}
Jia, D., Hermans, A., and Leibe, B. (2022).
\newblock {2D} vs. {3D} {LiDAR}-based {Person} {Detection} on {Mobile}
  {Robots}.
\newblock In {\em 2022 {IEEE}/{RSJ} {International} {Conference} on
  {Intelligent} {Robots} and {Systems} ({IROS})}, pages 3604--3611.

\bibitem[Lang et~al., 2019]{pointpillars}
Lang, A.~H., Vora, S., Caesar, H., Zhou, L., Yang, J., and Beijbom, O. (2019).
\newblock {PointPillars}: {Fast} {Encoders} for {Object} {Detection} {From}
  {Point} {Clouds}.
\newblock In {\em 2019 {IEEE}/{CVF} {Conference} on {Computer} {Vision} and
  {Pattern} {Recognition} ({CVPR})}, pages 12689--12697.

\bibitem[Le et~al., 2024]{le_jrdb-panotrack_2024}
Le, D.~T., Gou, C., Datta, S., Shi, H., Reid, I., Cai, J., and Rezatofighi, H.
  (2024).
\newblock {JRDB}-{PanoTrack}: {An} {Open}-{World} {Panoptic} {Segmentation} and
  {Tracking} {Robotic} {Dataset} in {Crowded} {Human} {Environments}.
\newblock In {\em 2024 {IEEE}/{CVF} {Conference} on {Computer} {Vision} and
  {Pattern} {Recognition} ({CVPR})}, pages 22325--22334.

\bibitem[Li et~al., 2023]{bevdepth}
Li, Y., Ge, Z., Yu, G., Yang, J., Wang, Z., Shi, Y., Sun, J., and Li, Z.
  (2023).
\newblock {BEVDepth}: {Acquisition} of {Reliable} {Depth} for {Multi}-{View}
  {3D} {Object} {Detection}.
\newblock {\em Proceedings of the AAAI Conference on Artificial Intelligence},
  37(2):1477--1485.

\bibitem[Linder et~al., 2021]{linder_cross-modal_2021}
Linder, T., Vaskevicius, N., Schirmer, R., and Arras, K.~O. (2021).
\newblock Cross-{Modal} {Analysis} of {Human} {Detection} for {Robotics}: {An}
  {Industrial} {Case} {Study}.
\newblock In {\em 2021 {IEEE}/{RSJ} {International} {Conference} on
  {Intelligent} {Robots} and {Systems} ({IROS})}, pages 971--978.

\bibitem[Liu et~al., 2023]{bevfusion}
Liu, Z., Tang, H., Amini, A., Yang, X., Mao, H., Rus, D.~L., and Han, S.
  (2023).
\newblock {BEVFusion}: {Multi}-{Task} {Multi}-{Sensor} {Fusion} with {Unified}
  {Bird}'s-{Eye} {View} {Representation}.
\newblock In {\em 2023 {IEEE} {International} {Conference} on {Robotics} and
  {Automation} ({ICRA})}, pages 2774--2781.

\bibitem[Martín-Martín et~al., 2023]{martin-martin_jrdb_2023}
Martín-Martín, R., Patel, M., Rezatofighi, H., Shenoi, A., Gwak, J., Frankel,
  E., Sadeghian, A., and Savarese, S. (2023).
\newblock {JRDB}: {A} {Dataset} and {Benchmark} of {Egocentric} {Robot}
  {Visual} {Perception} of {Humans} in {Built} {Environments}.
\newblock {\em IEEE Transactions on Pattern Analysis and Machine Intelligence},
  45(6):6748--6765.

\bibitem[Song et~al., 2024]{song_robustness-aware_2024}
Song, Z., Liu, L., Jia, F., Luo, Y., Jia, C., Zhang, G., Yang, L., and Wang, L.
  (2024).
\newblock Robustness-{Aware} {3D} {Object} {Detection} in {Autonomous}
  {Driving}: {A} {Review} and {Outlook}.
\newblock {\em IEEE Transactions on Intelligent Transportation Systems},
  25(11):15407--15436.

\bibitem[Yu et~al., 2023]{yu_benchmarking_2023}
Yu, K., Tao, T., Xie, H., Lin, Z., Liang, T., Wang, B., Chen, P., Hao, D.,
  Wang, Y., and Liang, X. (2023).
\newblock Benchmarking the {Robustness} of {LiDAR}-{Camera} {Fusion} for {3D}
  {Object} {Detection}.
\newblock In {\em 2023 {IEEE}/{CVF} {Conference} on {Computer} {Vision} and
  {Pattern} {Recognition} {Workshops} ({CVPRW})}, pages 3188--3198.

\end{thebibliography}

\end{document}